\documentclass[11pt,a4paper]{article}
\usepackage[hyperref]{eacl2021}
\usepackage{times}
\usepackage{latexsym}

\usepackage{multicol,multirow}
\usepackage{tabularx} 
\usepackage{graphicx}
\usepackage{tikz}
\usepackage{amsmath,amssymb}
\usepackage{booktabs}
\usepackage{arydshln}

\usepackage{mathrsfs}

\newcommand{\mtt}[1]{$\mathtt{#1}$}

\usepackage{microtype}

\aclfinalcopy 
\def\aclpaperid{1359} 

\title{{\sc GLaRA}: Graph-based Labeling Rule Augmentation for Weakly Supervised Named Entity Recognition}

  \author{Xinyan Zhao$^*$  \quad Haibo Ding$^\dagger$ \quad Zhe Feng$^\dagger$ \\
   $^*$School of Information, 
     University of Michigan, MI 48109, USA  \\
   $^\dagger$ Bosch Research and Technology Center, 
    Sunnyvale, CA 94085, USA  \\
   {\tt zhaoxy@umich.edu,  \{haibo.ding,zhe.feng2\}@us.bosch.com }\\
   }

\date{}

\begin{document}
\maketitle
\begin{abstract}


Instead of using expensive manual annotations, 
researchers have proposed to train named entity recognition (NER) systems using heuristic labeling rules.
However, devising labeling rules is challenging because it often requires a considerable amount of manual effort and domain expertise.
To alleviate this problem, we propose \textsc{GLaRA}, a graph-based labeling rule augmentation framework, to learn new labeling rules from unlabeled data.
We first create a graph with nodes representing candidate rules extracted from unlabeled data. 
Then, we design a new graph neural network to augment labeling rules by exploring the semantic relations between rules.
We finally apply the augmented rules on unlabeled data to generate weak labels and train a NER model using the weakly labeled data. 
We evaluate our method on three NER datasets and find that we can achieve an average improvement of +20\% F1 score over the best baseline when given a small set of seed rules.



\end{abstract}

\section{Introduction}
\label{sec:intro}


Named entity recognition (NER) models often need to be trained with many manual labels to perform well. 
Due to the high cost of manual annotations, collecting labeled data to train NER models is challenging for real-world applications. 
Recently, researchers have proposed to collect weak labels using heuristic rules, which are called labeling rules~\cite{bach2017learning,fries2017swellshark,ratner2020snorkel,safranchik2020weakly}. 
This kind of methods typically first ask domain experts to write labeling rules for a NER task, then use these manual rules to generate labeled data and train a NER model with the weakly labeled data. 
The advantage of these methods is that they do not require manual annotations. 
However, during our study, we find that writing labeling rules is also challenging for domain-specific tasks. 
Devising accurate rules often demands a significant amount of manual effort because it requires developers to have deep domain expertise and a thorough understanding of the target data.

\begin{figure}[t]
\centering
\includegraphics[trim={7.5cm, 4.3cm, 7.5cm, 1.5cm}, clip, width=\linewidth]{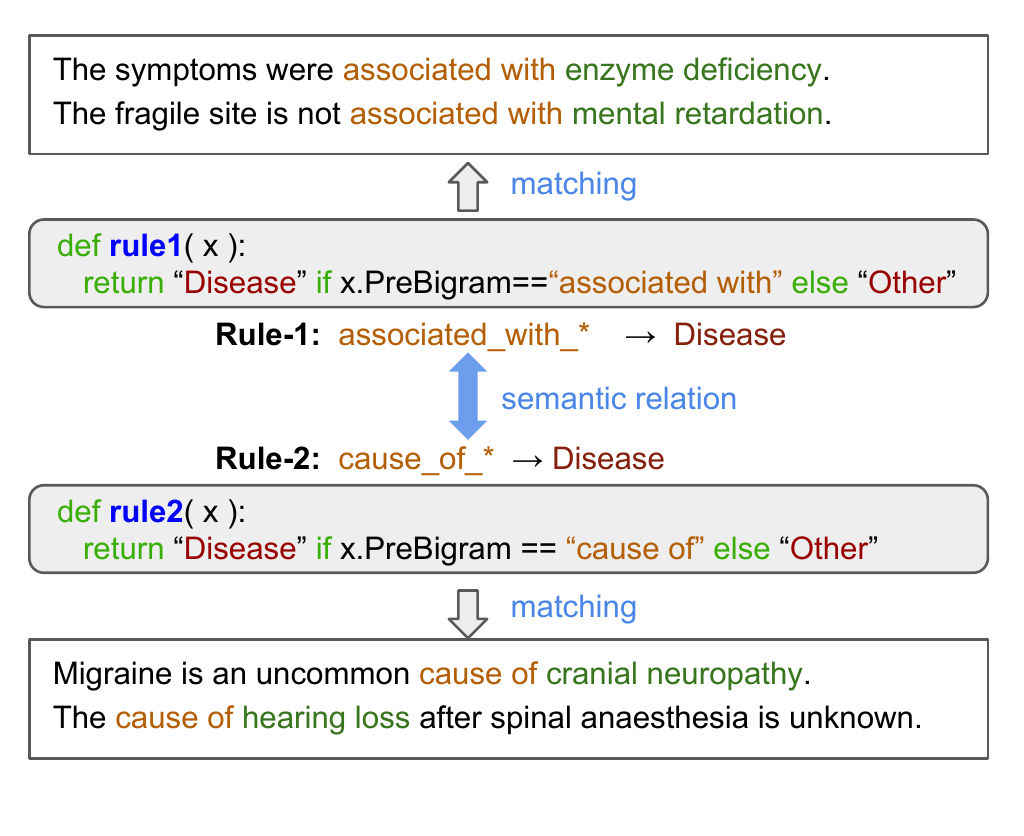}
\caption{Two examples of rules for recognizing \mtt{Diseases}. \mtt{PreBigram} means the 2 tokens on the left of a candidate entity.}
\label{fig:intuition-example}
\end{figure}

\begin{figure*}[t]
    \centering
     \includegraphics[trim={1cm 2cm 1cm 2cm},clip,width=\textwidth]{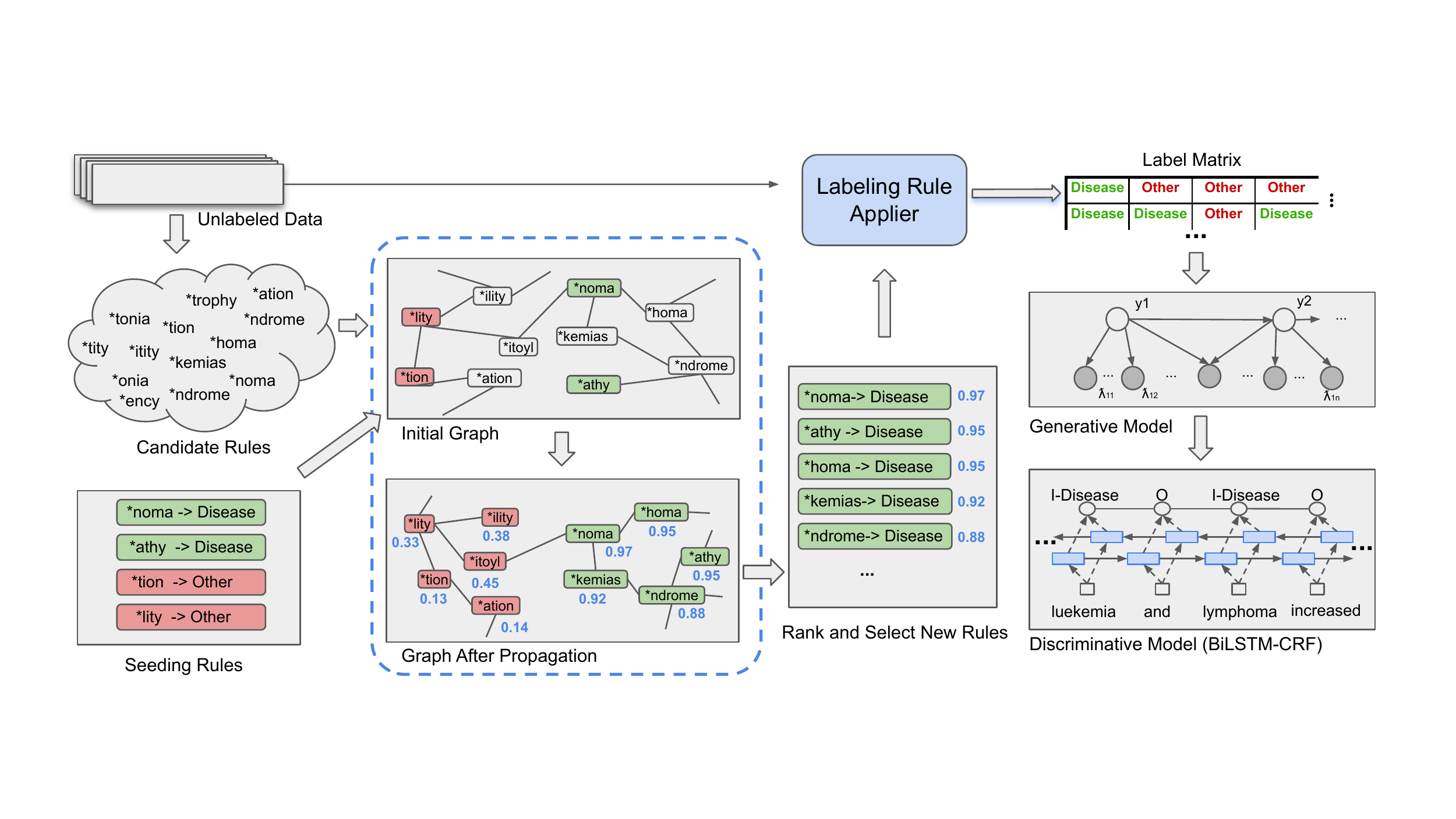}
    \caption{An example workflow of \textsc{GLaRA} framework that using \mtt{suffix~rules} to recognize \mtt{Diseases}. 
    Given unlabeled data and seeding rules, (1) we first extract candidate rules from unlabeled data, (2) then build a graph of rules and learn new rules by propagating the labeling information from seeding rules to other rules. (3) Next we apply selected rules on unlabeled data and obtain a label matrix. (4) Finally, we estimate the noisy labels using a generative model, and (5) train a final NER model with noisy labels.
    ``\mtt{*noma\rightarrow Disease}'' denotes that if a word's suffix is ``\mtt{noma}'' then it will be labeled as \mtt{Disease}.
    }
    \label{fig:workflow}
\end{figure*}

To alleviate manual effort on writing labeling rules, we propose \textsc{GLaRA}, 
a \underline{\textbf{G}}raph-based 
\underline{\textbf{La}}beling 
\underline{\textbf{R}}ule  
\underline{\textbf{A}}ugmentation framework to automatically learn new rules 
from unlabeled data with a handful of seed rules. 
Our work is motivated by the intuition that if two rules can accurately label the same type of entities, then they are semantically related via the entities matched by them.
Therefore, we can acquire new labeling rules based on their semantic relatedness with the rules that we have already known. 
For example, Figure \ref{fig:intuition-example} shows two example rules for labeling \mtt{Disease} entities. 
If we know that \mtt{rule1} ``\mtt{associated~with~*}$\rightarrow$\mtt{Disease}'' is an accurate rule for labeling diseases, and \mtt{rule1} is semantically related to \mtt{rule2} ``\mtt{cause~of~*}$\rightarrow$\mtt{Disease}'', 
then we can derive that \mtt{rule2} can be another accurate rule for labeling diseases. 


To augment labeling rules, 
we first define six types of rules and extract all possible rules from unlabeled data as candidate rules. 
Then, for each rule type, we build a graph by connecting rules of this type based on their semantic similarities. 
In our work, we compute a rule's embedding as the average of contextual embeddings of entities matched by the rule, and 
compute the similarities between rules using their embeddings.
We learn new rules using a graph neural network model from a small set of seed rules. 
Next, we train a discriminative NER model using the weak labels generated by both the seeding and learned rules. 
To obtain weak training labels, 
we first obtain a label matrix by applying all the augmented rules on each token in unlabeled data. 
Then, we estimate the labels of unlabeled instances using the LinkedHMM model \cite{safranchik2020weakly}.
We evaluate our framework on three datasets. 
In our experiments,
we first show that our method can achieve better results than baselines when abundant rules are available. We also demonstrate that we can achieve an average improvement of +20\% F1 when only a small set of rules are available. 

We summarize our major contributions as:
\begin{itemize}
    \item We propose a new Graph-based Rule Labeling Rule Augmentation ({\sc GLaRA})\footnote{The code is available at~\url{https://github.com/zhaoxy92/GLaRA}.} framework, which can effectively learn new labeling rules from unlabeled data automatically. 
    \item We propose a new graph neural network to estimate rules' labeling confidence with a new class distance-based loss function.
    \item We define six types of labeling rules, which have been proven to be effective on three named entity recognition tasks.
\end{itemize}

\section{The \textsc{GLaRA} Framework}
\label{sec:pattern}


Our goal is to build a NER system with a small set of manually selected seeding rules and unlabeled data. 
Our key idea is to first augment labeling rules using graph neural networks, based on the hypothesis that semantically similar rules should have similar abilities to recognize entities. 
Then, we train a NER model using the weak training data labeled by the augmented rules. 


\paragraph{Overview}
Figure \ref{fig:workflow} shows an example workflow of \textsc{GLaRA} framework using \mtt{suffix} rules to recognize \mtt{Disease} entities. 
Our framework consists of five major components.
(1) \emph{Rule extractor}: We define six types of rules and extract all possible rules from unlabeled data as candidate rules. 
(2) \emph{Rule augmentation}: For each rule type, 
we first build a graph of rules by connecting rules based on their semantic similarities. 
Given a small set of manual seeding rules, 
we learn new rules by propagating the labeling confidence from seeding rules to other rules. 
(3) \emph{Rule Applier}: We obtain a label matrix by applying all the augmented rules on each token in the unlabeled data. 
(4) \emph{Generative Model}: We estimate the labels of unlabeled instances using a generative model.
(5) \emph{Discriminative Model}: We train a final discriminative NER model using the weak labels produced by the generative model.

\subsection{Candidate Rule Extraction}
\label{sec:rule-extraction}
As demonstrated in previous work on NER \cite{zhou2002named}, lexical and contextual clues are strong indicators for entity recognition.
Therefore, we define and extract six types of rules: \mtt{SurfaceForm}, \mtt{Prefix}, \mtt{Suffix}, \mtt{PreNgram}, \mtt{PostNgram}, and \mtt{DependencyRule} to recognize entities by considering their lexical, contextual, and syntax information.

Given an unlabeled sentence, we first extract all noun phrases (NPs) using a set of Part-of-Speech (POS) patterns, as candidate entity mentions.
The POS patterns include ``JJ? NN+'' (JJ denotes an adjective, and NN denotes a noun) and top $15$ most frequent POS patterns of the entity mentions in the development sets.
Then, we extract all six types of rules from the unlabeled data as candidate rules for each candidate entity mention.
Specifically, we extract the surface form of each candidate entity mention as a \mtt{SufraceForm} rule. 
If the mention is a single token, we extract its first and last $m$ characters as \mtt{Prefix} and \mtt{Suffix} rules, respectively. 
For \mtt{PreNgram} rule, we extract leading $n$ words of a candidate entity as \mtt{inclusive~PreNgram} rule; meanwhile, we also extract $n$ words on the left of candidate as \mtt{exclusive~PreNgram} rule. \mtt{PostNgram} rules are created similarly from the right context. 
Also, for each multi-token entity candidate, we first extract the dependency relations of the first token and the second last token, respectively, and then each dependency is combined with the last token as \mtt{Dependency} rules of this mention.
We treat $m$ and $n$ as hyperparameters in our work.
%
%
%
%

\begin{figure}[t]
    \centering
     \includegraphics[trim={7.6cm 1.5cm 8cm 0.7cm},clip,width=\linewidth]{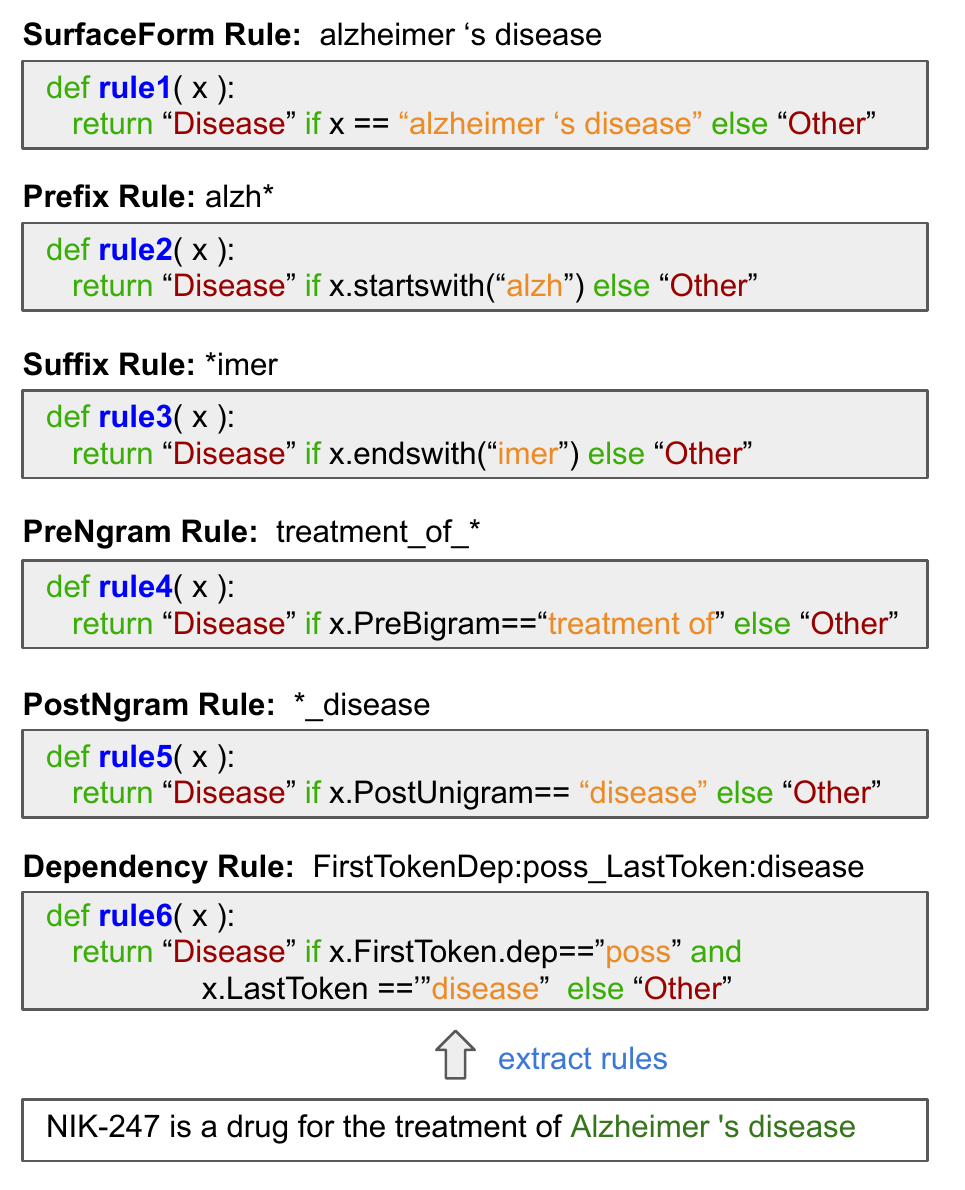}
    \caption{ Example rules extracted for an entity mention, and how these rules will be used to label entities. 
    }
    \label{fig:rule-examples}
\end{figure}

Figure~\ref{fig:rule-examples} show some example rules that extracted for the candidate entity mention ``{\it Alzheimer 's disease}'' 
and how they are used for labeling entities. 


\subsection{Labeling Rule Augmentation}

We aim to learn new labeling rules from seed rules by exploiting the semantic relations between rules. 
Specifically, we first build a graph with rules as nodes. 
Then, we initialize the graph with both manually selected positive and negative seeding rules. 
Positive seeding rules are those that can be used to predict a target entity type.
Negative ones are used to predict instances of the ``{\it Other}'' class. 
Next, we estimate the representations of rules by optimizing a graph neural network model.
Finally, we compute the distances of a rule to the centroids of positive and negative seeding rules, respectively,
and then select rules close to positive centroid as new labeling rules.


\paragraph{Graph of Rules}
For each type rules,
we create a graph $G=(\mathcal{V}_{u}, \mathcal{V}_{s}^{pos}, \mathcal{V}_{s}^{neg}, \mathcal{A})$ with this type of rules as nodes, where $\mathcal{V}_{u}$ denotes the candidate rules extracted in the previous step, $\mathcal{V}_{s}^{pos}$ and $\mathcal{V}_{s}^{neg}$ denotes the positive and negative seeding rules, respectively.  
$\mathcal{A}$ is the adjacency matrix of nodes. 
In our graph, each node (i.e., rule) is connected with the top 10 semantically similar nodes. The similarity between two rules is computed as the cosine similarity using their embeddings.

\paragraph{Rule Embeddings} We estimate semantic relatedness between rules with their embeddings, which are computed using pre-trained contextual embedding models. Specifically, we first apply the pre-trained  ELMo~\cite{peters2018deep} model on all unlabeled sentences to obtain contextual embeddings of candidate entity mentions. 
Then, we compute the embedding of a rule as the average of the embeddings of all the candidate entities that can be matched by this rule. 
For example, the embedding of the prefix rule ``*noma'' is calculated as the average of the embeddings of all candidate mentions that ends with ``noma''.


\paragraph{Graph Propagation Model}
Given a graph of rules and seeding rules, we formulate the problem of learning new labeling rules (i.e., positive rules) as a graph-based semi-supervised node classification task that aims to classify candidate rules (which could be treated as unlabeled nodes in the graph) as positive or negative.

Based on the intuition that semantically similar rules should predict entity labels similarly. 
We propose a graph neural network model to propagate labeling information from seeding nodes to other nodes based on the recent work on Graph Attention Network \cite{velivckovic2017graph}.
%
%
Specifically, 
given the input embedding $h_i$ of node $i$ and its neighbors $\mathcal{N}_i$, we first compute an attention weight 
for each connected pair ($i,j$) as,
\begin{equation}
\alpha_{ij} = \frac{exp(f(A^{T}[Wh_{i}, Wh_{j}])))}{\sum_{k\in N_{i}} exp(f(A^{T}[Wh_{i},Wh_{k}]))} 
\end{equation}
where $W$ is a parameter matrix, and $f$ is the LeakyReLU activation function.
We then re-compute the embedding of $i$:
\begin{equation}
\label{eq2:new-state}
 h_{i}^{*} = \alpha_{i,i}Wh_{i} + \sum_{j\in N_{i}} Wh_{j}
\end{equation}
To keep model stable, we apply multi-head attention 
mechanism to obtain $K$ attentional states for 
each node, the average of which is used as final 
node representation, i.e., $h_{i}^{*T} = \frac{1}{K} \sum_{k}h_{i}^{kT}$. 

The objective of our model is defined as follows:
\begin{equation}
    \mathcal{L}_{total} = \mathcal{L}_{sup} +  \mathcal{L}_{reg} +  \mathcal{L}_{dist}  
\label{eq3:loss}
\end{equation}
where 
$ \mathcal{L}_{sup} = -(y_{i} \log(p_{i})) + (1-y_{i}) \log(1-p_i)  $

\hspace{0.6cm}$ \mathcal{L}_{reg} = \sum_{i,j\in N_{i}} \|h_{i}-h_{j}\|_{2}  $

\hspace{0.6cm}$ \mathcal{L}_{dist} = dist(h_{pos}, h_{neg})$

\noindent
where $\mathcal{L}_{sup}$ is the supervision loss computed on both positive and negative seeding rule nodes, 
$\mathcal{L}_{reg}$ is the regularization that encourages connected nodes to share similar representations, 
and $\mathcal{L}_{dist}$ aims to maximize the distance between positive
and negative seeding nodes. $dist(\cdot)$ computes the cosine similarity between the centroids of the positive and negative seeds.
$p_i$ is the probability of a node being classified as positive, and $h_{pos}$ and $h_{neg}$ 
are the average embeddings of positive and negative nodes, respectively. 

When the learning process is finished, each rule is associated with  a new embedding representation, i.e.,  $h_{i}^{*T}$. 
For each rule, we first compute its cosine distances to the centroids of positive and negative seeding nodes using their embeddings, respectively. 
Then, we rank all rules by the difference of two distances and select the top $M$ rules closest to the centroid of the positive seeding rules as new labeling rules.


\subsection{Generative Model for Label Estimation}
After the rule learning process, we apply both the newly learned rules and the seeding rules on unlabeled data to produce a matrix of labels.\footnote{In our work, we also applied the linking rules developed by \cite{safranchik2020weakly} on unlabeled data, so the label matrix also contains the results from linking rules.}
Since one token can be potentially matched by several different rules, the resulting labels can have conflicts. 
Therefore, we use the LinkedHMM \cite{safranchik2020weakly} model to combine these labels into one label for each token. 
Briefly, the main idea of LinkHMM  is to treat the true label of a token as a latent random variable and estimate its value by relating it to the label outputs from different labeling rules. 
The estimated labels can be used to train final discriminative NER models.


\subsection{Discriminative NER Model}
After the training of the generative model (i.e., the LinkedHMM in our work) is completed, each token in the unlabeled data is associated with a weak label.
Each weak label is a probability distribution over all entity classes, which can be used to train a discriminative NER model. 
One advantage of training a discriminative NER model 
is that it can use other token features while 
the generative model can only use labeling rules' outputs as inputs. 
Therefore, even if a token is not matched by any labeling rules, it can still be predicted correctly by the discriminative model.


In our work, we use BiLSTM-CRF~\cite{huang2015bidirectional} as our discriminative model. 
The model first uses BiLSTM to generate a state representation for each token in a sequence. 
The CRF layer then predicts each token by maximizing the expected likelihood of the entire sequence based on the estimated labels.

\section{Experimental Setup}
In this section, we present the details of our experimental setup. 
First, we evaluate our method on three NER datasets, which 
we refer to \mtt{NCBI}, \mtt{BC5CDR}, and \mtt{LaptopReview}. 
Then, we describe the baseline methods compared in our experiments. 
We also give a detailed description of the seed rules used in our experiments. 


\subsection{Datasets}
We evaluated our method on three datasets. Details of each dataset are described below.

\mtt{NCBI}~\cite{dougan2014ncbi} contains 793 PubMed abstracts with 6,892 \texttt{Disease} mentions and is split into 593 train, 100 dev and 100 test data.

\mtt{BC5CDR}~\cite{li2016biocreative} has 1,500 PubMed articles with 5,818 \texttt{Disease} and 3,116 \texttt{Chemical} mentions. It is split into train, dev, and test sets with 500 articles each. 

\mtt{LaptopReview}~\cite{pontiki2016semeval} contains sentences regarding laptop \texttt{AspectTerms}, including 3,048 training and 800 test sentences. Following~\cite{safranchik2020weakly}, we hold out $20\%$ of training data as dev set.

Note that we use all training data as our unlabeled data by removing the manual annotations.



\subsection{Compared Methods}
We compare our method with state-of-the-art weakly supervised methods using dictionaries or heuristic rules as supervision. 
In this section, we briefly describe these baseline methods.
\paragraph{AutoNER}~\cite{shang2018learning} is a distantly supervised method, which automatically builds a neural named entity recognition model using dictionaries as weak supervision. 
\paragraph{Snorkel}~\cite{ratner2020snorkel} is a general machine learning framework that can train classifiers using heuristic rules. By default, it uses a Naive Bayes generative model to denoise labeling rules by predicting each token's label independently.  
\paragraph{SwellShark}~\cite{fries2017swellshark} is an extension of Snorkel that was developed for biomedical NER. Same as Snorkel, it uses a naive Bayes generative model to denoise manual labeling rules.  
It also requires a special entity candidate generator to detect entity spans accurately before predicting their entity labels. 
In our experiments, we report both results using simple noun phrases as candidates, and that generated using extra expert effort.  
\paragraph{LinkedHMM}~\cite{safranchik2020weakly} is a framework for training sequence tagging models using weak supervision from manual rules. 
Besides using labeling rules, it can also use linking rules that indicate whether two consecutive tokens have the same label.



\begin{table}[h]
\small
\begin{tabular}{p{12mm}p{2mm}p{2mm}p{3mm}p{2mm}p{3mm}p{2mm}p{2mm}p{2mm}}
\toprule
\textbf{RuleType} & \multicolumn{2}{l}{NCBI} & \multicolumn{2}{l}{\begin{tabular}[c]{@{}l@{}}BC5CDR\\ (Disease)\end{tabular}} & \multicolumn{2}{l}{\begin{tabular}[c]{@{}l@{}}BC5CDR\\ (Chem)\end{tabular}} & \multicolumn{2}{l}{Laptop} \\& S & M& S& M& S & M & S & M \\\hline
\multicolumn{9}{c}{Positive seeding rules}\\
Surface  & 875 & 0& 632K    & 0  & 1.5M & 0  & 274 & 0  \\
Suffix     & 12 & 5  & 14  & 2    & 7  & 9  & 0   & 5  \\
Prefix     & 0 & 7 & 2  & 13 & 0 & 8  & 0  & 5 \\
PreNgram$^{\dagger}$   & 5 & 9  & 0& 9   & 1  & 8  & 6 & 5 \\
PostNgram$^{\dagger}$  & 4 & 6   & 5 & 10  & 2  & 9 & 0  & 5   \\
DepRule & 5  & 8  & 5   & 7   & 0  & 6  & 0 & 4 \\\hline
\multicolumn{9}{c}{Negative seeding rules}\\
Surface & 34 & 0 & 34 & 6 & 34 & 0 & 34 & 3\\
Suffix & 0 & 10 & 0 & 18 & 0 & 22 & 0 & 5\\
Prefix & 0 & 10 & 0 & 12 & 0 & 8 & 0 & 8\\
PreNgram$^{\dagger}$ & 0 & 15 & 0 & 18 & 0 & 16 & 0 & 9\\
PostNgram$^{\dagger}$ & 0 & 10 & 0 & 19 & 0 & 10 & 0 & 7\\
DepRule & 0 & 8 & 0 & 13 & 0 & 5 & 0 & 4\\

\bottomrule
\end{tabular}
\caption{Number of positive and negative seed rules for each dataset. \mtt{S} denotes the number of seed rules by~\cite{safranchik2020weakly} and \mtt{M} denotes the rules we manually selected. $\dagger$ denotes that corresponding rule type includes both \mtt{inclusive} and \mtt{exclusive} rules.}
\label{tab:seed-num}
\end{table}




\setlength\dashlinedash{1pt}
\setlength\dashlinegap{1pt}
\begin{table*}[ht]
\centering
\small
\begin{tabular}{llllllllllll}
\toprule
\multirow{2}{*}{ {\bf Method}} & 
\multirow{2}{*}{ {\bf Human Effort}} 
& \multicolumn{3}{c}{{\bf NCBI}} 
        & \multicolumn{3}{c}{{\bf BC5CDR}} & \multicolumn{3}{c}{{\bf LaptopReview}} \\
& &  {\bf P } & {\bf R} & {\bf F1} & {\bf P} & {\bf R} & {\bf F1} & {\bf P} & {\bf R} & {\bf F1} \\
\midrule
Supervised & Full Annotations & 85.2 & 89.2 & 87.2 & 87.2 & 88.0 & 87.5 & 83.5 & 82.2 & 82.9 \\
\midrule
Snorkel    & Saf.'s Manual Rules & - & - & 68.7 & - & - & 83.2 & - & - & 60.0\\ \hdashline
SwellShark & Fries' Manual Rules & 64.7 & 69.7 & 67.1 & 85.0 & 83.5 & 84.2 & - & - & - \\
    \multicolumn{2}{r}{+Special Candidate Generator} & 81.6 & 80.1 & {\bf 80.8} & 86.1 & 82.4 & 84.2 & - & - & - \\ \hdashline
AutoNER  & Dictionaries & 79.4 & 72.0 & 75.5 & 89.0 & 81.0 & 84.8 & 72.3& 59.8 & 65.4\\ \hdashline
LinkedHMM  & Saf.'s Manual Rules & - & - & 78.7 & - & - & 85.9 & - & - & 68.2 \\
       & Our Seed Rules & 89.4& 70.7 & 79.0 & 88.0& 84.5 & 86.2 & 82.7 & 59.8& 69.4 \\ \hline
{\sc GLaRA}  & Our Seed Rules & 89.9 & 73.2 & 80.2{\tiny$\pm.2$} & 88.2 & 84.6 & \textbf{86.3}{\tiny$\pm.3$} & 82.4 & 64.2 & {\bf 72.2}{\tiny$\pm1.2$} \\ 
\bottomrule
\end{tabular}
\caption{Micro F1 performance of baselines and our method on test sets. 
Our results are the mean and std across 5 random runs.
\mtt{Saf.'s~Manual~Rules} are the rules developped by \citet{safranchik2020weakly}, and \mtt{Fries'~Manual~Rules} are those 
used in \cite{fries2017swellshark}.
}
\label{tab:main_results}
\end{table*}

\subsection{Seed Rules}
\label{sec:manual-rules}
In our experiments, we used all the labeling rules\footnote{In our experiments, we used the linked rules from \cite{safranchik2020weakly} to obtain weak labels, which are another type of rules that can be used to vote if the two consecutive tokens have the same label, to train our LinkedHMM model. However, our work only focused on augmenting labeling rules, and did not augment linking rules.}
developed by \citet{safranchik2020weakly} as part of positive seeding rules. 
Besides, we manually 
selected another small set of labeling rules as our input because: 
(1) we defined six types of rules as described in Section \ref{sec:rule-extraction}, but some types of rules were not used in \citet{safranchik2020weakly} (e.g., prefix rules are not used in NCBI), 
(2) our method requires negative seed rules, which are used to identify terms that are not entities, to initiate its learning process. 
To automatically learn new labeling rules, we use both labeling rules from \cite{safranchik2020weakly} and our manually selected rules as seeding rules. 
Numbers of both positive and negative seeding rules used in our experiments are shown in Table \ref{tab:seed-num}.\footnote{Note that in previous work a whole lexicon or dictionary is counted as one rule. However, in our work we count each term in a lexicon or dictionary as one surface form rule.} 
Our manually selected seed rules are included in Appendix C.

\subsection{Implementation Details}
\label{sec:implement}

In our experiments, we create a graph for each type of rules for each dataset and learn new rules independently with the same setup. \mtt{Prefix} and \mtt{Suffix} candidate rules are generated by considering the first and last 3 to 6 characters, and \mtt{PreNgram} and \mtt{PostNgram} candidate rules are extracted with the windows of 1 to 3 tokens.

We use a two-layer graph attention network to train our graph model.
After training, we select $M$ new rules for each type of rules, where the value of $M$ is searched between 20 and 500 on dev sets. 

For different datasets, our discriminative NER models used different pre-trained contextual models. 
Since NCBI and BC5CDR datasets are in the biomedical domain, we finetuned our NER model on the pretrained SciBERT embedding~\cite{beltagy2019scibert}, while for LaptopReview data, the NER model is finetuned on the pretrained BERT$_{base}$~\cite{devlin2018bert} embedding.
More details are provided in Appendix A.

\section{Experimental Results}

In this section, we first compare our method with state of the art 
methods using all available manual rules. 
Then, we evaluate our method under scenarios when only limited seeding rules are available, 
which are very common in real-world applications. 
Next, we conduct an ablation study to investigate the effectiveness of different types of rules and our newly proposed loss function based on centroid distance.
Finally, we also perform a quality analysis of the automatically learned labeling rules.

\subsection{Results with Abundant Seeding Rules}
\label{sec:result-abundant}
In this subsection, we report both the result of our {\em generative} model with augmented rules in Table \ref{tab:main_results}  and the result of our {\em discriminative} NER model that is trained using weak labels from the generative model in Table \ref{tab:result-lstm}.

Table \ref{tab:main_results} shows the performance of our generative model with augmented rules
and baseline generative models\footnote{Some scores were reported in \cite{safranchik2020weakly}}. 
The {\em Supervised} line is the performance of the LinkedHMM model trained on fully labeled training data. 
The results show that our method with augmented labeling rules performed best on BC5CDR and LaptopReview datasets. We also achieved a comparative F1 score of 80.2 on NCBI with SwellShark (F1 80.8). Note that SwellShark used lots of manual effort from experts to carefully tune a particular candidate generator for a given dataset.
We also notice that, with augmented rules, our method 
outperforms LinkedHMM by an average of 2.0 F1 points, which 
demonstrates the effectiveness of augmented rules.

\begin{table}[h]
\small
\begin{tabular}{lp{10mm}p{10mm}p{15mm}} 
\toprule
{\bf WeakSupervision} & NCBI & BC5CDR &      Laptop\\
      \midrule

Snorkel & 73.4{\tiny$\pm1.7$} & 82.2{\tiny$\pm.5$} & 63.5{\tiny$\pm1.7$} \\
LinkedHMM  &  79.0{\tiny$\pm.4$} &  83.0{\tiny$\pm.2$}  & 69.0{\tiny$\pm1.1$}  \\
{\sc GLaRA} &  {\bf 80.8}{\tiny$\pm.3$} & {\bf 83.5}{\tiny$\pm.6$} & {\bf 72.3}{\tiny$\pm1.0$}  \\
      
\bottomrule
\end{tabular}
\caption{Results of discriminative models using weak labels generated by different methods. Our results are the average of five runs.}
\label{tab:result-lstm}
\end{table}

\begin{figure*}[t]
    \centering
    \includegraphics[width=\textwidth]{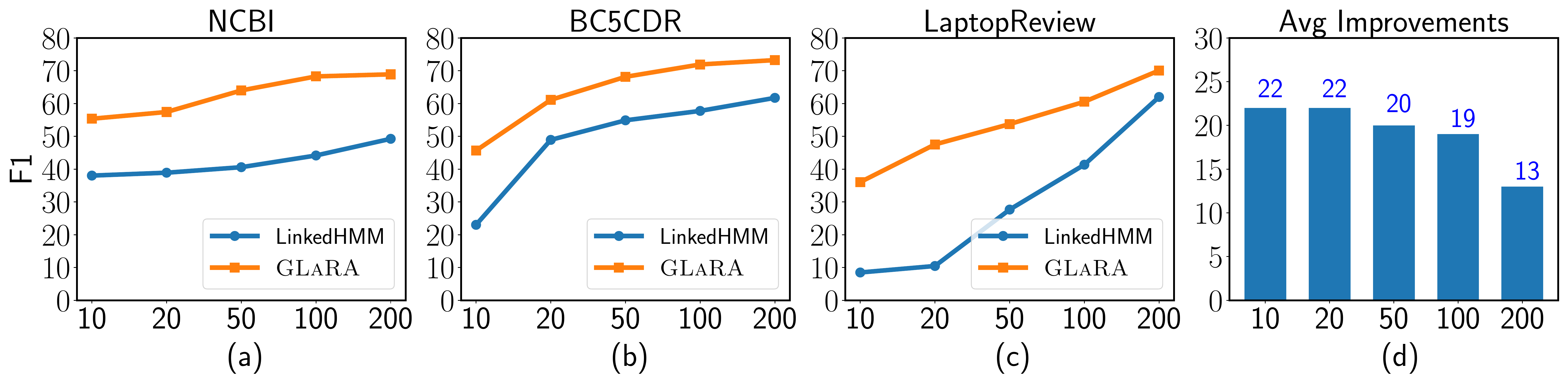}
    \caption{Results of LinkedHMM with only seeding rules and augmented rules by our {\sc GLaRA} method. The bottom numbers are the numbers of seeding rules used.}
    \label{fig:performance-diff-seed}
\end{figure*}

Table \ref{tab:result-lstm} shows the performance of our discriminative NER model (BiLSTM-CRF) trained on weak labels from our \textsc{GLaRA} with augmented rules and baseline discriminative models using weak labels from LinkedHMM and Snorkel without augmented rules \cite{safranchik2020weakly}. 
The results show that our method achieved a +1.9 F1 point improvement over the second-best system. We also notice that our discriminative model performs slightly better on \mtt{NCBI} and \mtt{Laptop}, but worse on \mtt{BC5CDR} than the generative model, which is consistent with that from \cite{safranchik2020weakly}.

\subsection{Results with Limited Seeding Rules}

Though weakly supervised state-of-the-art methods reported in Table \ref{tab:main_results} achieved relatively good performance, they require a significant amount of manual effort from domain experts for designing and tuning labeling rules. 
Well-performing methods that require less manual effort are often more desirable.
Therefore, we also evaluate our method with little manual effort (i.e., limited seeding rules). 
We conducted experiments by randomly select at most $k$ rules for 
each rule type from our seed rules (Section \ref{sec:manual-rules}), where $k \in$ \{10, 20, 50, 100, 200\}. 
When a rule type has less than $k$ seeding rules, we use all of them. 
Figure \ref{fig:performance-diff-seed} shows the performance of the LinkedHMM model using only seeding rules and our {\sc GLaRA} method with augmented rules on three datasets. 
Figure~\ref{fig:performance-diff-seed}(d) shows that, 
with automatically learned rules, our method can 
obtain average F1 gains of +22 points when using 10 seeding rules 
and +13 points when there are 200 seeding rules.







\subsection{Impact of Different Types of Rules}

To investigate the effectiveness of different types of labeling rules, we conducted ablation experiments to evaluate our generative model with augmented rules (i.e., {\sc GLaRA}) by adding each type of rules cumulatively.
Results are shown in Table~\ref{tab:ablation}.
Empty cells denote that the corresponding type of rules are not used. 
Results show that \mtt{Prefix} rules are most effective on the Laptop dataset, producing +1.5 F1 point improvement. 
All other rules except \mtt{PostNgram} can improve the performance by +0.4 to +0.6 F1 points.




\begin{table}[h]
    \centering
\begin{tabular}{lccc  c}
\toprule
          & NCBI & CDR & Laptop & $\Delta$ \\ 
         \midrule 
LinkedHMM & 78.7 & 85.9 & 68.2 &  \\ \hline
\textsc{GLaRA}&  &  &  & \\
~~+DepRule   & 79.3 & -&-  & +0.6\\
~~+PostNgram & - & 86.0 &- & +0.1 \\
~~+PreNgram  &  79.6 & 86.2 & 69.3 & +0.5\\
~~+Prefix    & - &- & 70.8 & +1.5\\
~~+Suffix    & 79.8 & 86.3 & 71.7 & +0.4\\
~~+Surface   & 80.2  & - & 72.2 & +0.5\\ 
    \bottomrule
    \end{tabular}
    \caption{Impact of each type of rules used in our generative model. $\Delta$ denotes the average improvement achieved by adding rules cumulatively.}
    \label{tab:ablation}
\end{table}


\begin{table}[b]
\small
\centering
\begin{tabular}{lccc}
\toprule
{\sc GLaRA }      & NCBI &       BC5CDR &       LaptopReview  \\
      \midrule
 w/ $\mathcal{L}_{dist}$  & 80.2{\tiny$\pm$.2} & 86.3{\tiny$\pm$.3} &  72.2{\tiny$\pm$1.2} \\ 
w/o $\mathcal{L}_{dist}$ &  79.3{\tiny$\pm$1.2} &  86.1{\tiny$\pm$.1} &  69.7{\tiny$\pm$.8}\\
\bottomrule
\end{tabular}
\caption{Results of our generative model using augmented rules learned by our graph neural network model with and without distance loss. }
\label{tab:result-dist-loss}
\end{table}

\begin{table*}[h]
\centering
\begin{tabular}{ll}
\toprule
\textbf{RuleType} & \textbf{Top 5 learned rules}  \\\hline
\mtt{Surface}    & {neurologic\_disease}, {spherocytic\_hemolytic\_anemia}, 
            {episodic\_hemolytic\_anemia}, \\
           & {choroid\_plexus\_carcinoma}, 
            {sporadic\_human\_renal\_cell\_carcinoma}\\\hline 
\mtt{Suffix}     & {*ioma}, {*ophy}, {*ndrome}, {*umonia}, {*phoma}\\\hline
\mtt{Prefix}     & {mesot*}, {mesoth*}, {prost*}, {dyst*}, {scler*}\\\hline
\mtt{PreNgram}   & {stage\_ii\_*}, {breast\_/\_ovarian\_*},
           {heterogeneity\_/\_in\_*}, {a\_syndromal\_*}, \\
           & {pathology\_of\_*} \\\hline
\mtt{PostNgram}  & {*\_hepatitis}, {*\_cardiomyopathy}, {*\_carcinoma},  {*\_colitis}, {*\_lymphoma}\\\hline
\mtt{DepRule}    & {FirstTokenDep:amod\_LastToken:syndrome}\\
           & {FirstTokenDep:compound\_LastToken:sarcoma}\\
           & {SecondLastTokenDep:amod\_LastToken:dystrophy}\\
           & {SecondLastTokenDep:conj\_LastToken:dystrophy}\\
           & {SecondLastTokenDep:nmod\_LastToken:syndrome}\\
\bottomrule
\end{tabular}
\caption{Examples of learned new rules for recognizing diseases on NCBI data. ``\_'' denotes the space character. }
\label{tab:rule-example}
\end{table*}

\subsection{Effectiveness of Distance Loss $\mathcal{L}_{dist}$}

In this work, we proposed a new graph neural network model with a new loss function, i.e., $\mathcal{L}_{dist}$ in Eq.~\ref{eq3:loss}. This loss function measures the distance between the centroids of positive and negative seeding rules by computing their cosine similarity.
The motivation is that the model should keep positive rules distant from negative ones during the learning process. 
Table \ref{tab:result-dist-loss} shows the performance of our generative model with and without the distance loss. We find that our method can obtain an average F1 gain of 1.2 points across three datasets with this new loss function.

\subsection{Quality Analysis of Learned New Rules}

In Table~\ref{tab:rule-example}, we present top $5$ learned rules for each rule type that we automatically learned on the NCBI data set. Each rule is formatted in the way described in Section~\ref{sec:rule-extraction}. 
%

Table \ref{tab:rule-analysis} shows the analysis results of the number of new rules (\#Rule) and their accuracy (ACC) on the dev sets,
which were learned during the experiments presented in Section~\ref{sec:result-abundant}.
Results show that all the learned rules have accuracy $\geq$64\% with 
an average of 76\%, which justifies the quality of these new rules.

\begin{table}[h]
    \centering
    \begin{tabular}{llll}
    \toprule
    \textbf{RuleType} & \textbf{Dataset} &  \textbf{\#Rule} & \textbf{ACC}    \\
    \midrule
    \multirow{2}{*}{Surface} &  NCBI   & 301 & 78\%   \\
                             &  Laptop & 154 & 69\%   \\ \hdashline
    \multirow{3}{*}{Suffix}  & NCBI    & 64 & 100\%  \\
                             & CDR(Dis)& 138 & 90\%  \\
                             & Laptop  & 25 & 66\%   \\ \hdashline
    \multirow{1}{*}{Prefix}  & Laptop  & 16 & 78\%   \\ \hdashline
    \multirow{3}{*}{PreNgram}& NCBI    & 156 & 65\%  \\
                             & CDR(Dis)& 19 & 64\%   \\
                             & Laptop  & 17 & 71\%   \\
   \multirow{1}{*}{PostNgram}&CDR(Chem)& 37 & 82\%   \\ \hdashline
    \multirow{1}{*}{DepRule} & NCBI    & 47 & 70\%  \\ 
         \bottomrule
    \end{tabular}
    \caption{Quality analysis of learned new rules. }
    \label{tab:rule-analysis}
\end{table}


Besides, we also manually analyzed why some learned rules are not helpful to improve recognition performance. 
First, some of the learned rules can be overlapping. For example, both ``\mtt{*demia}$\rightarrow$\mtt{Disease}'' and 
``\mtt{*edemia}$\rightarrow$\mtt{Disease}'' can be used to recognize \mtt{Diseases}. However, if \mtt{*demia} is learned first, then learning \mtt{*edemia} rule will not help because the entities matched by these two rules have large overlaps. 
Second, some of the rules learned from unlabeled data may not be applied to testing data due to the mismatch between two datasets. 
For example, though the PostNgram rule ``\mtt{*\_dystonia}$\rightarrow$\mtt{Disease}'' is an accurate rule to label diseases such as ``oromandibular dystonia'' and ``responsive dystonia''. However, we do not find any matches on test data. 



\section{Related Work}
\label{sec:related-work}

Training reliable NER systems usually requires large annotation efforts, which is often considered expensive and impractical in certain domains. Therefore, previous studies have been trying to reduce the manual efforts required for annotation while producing comparable performance in NER tasks by utilizing manually designed rules that are cheap and accurate. 

Studies have shown that human-defined rules (i.e dictionaries) can greatly aid NER tasks, especially in the domains where the identification of entities needs domain knowledge~\cite{cohen2004exploiting,savova2010mayo,xu2010medex,eftimov2017rule}, and they can also be used to distantly create labeled data set to benefit machine learning models~\cite{mann2010generalized,neelakantan2015learning,giannakopoulos2017unsupervised}. 

While distantly labeled data sets can be created at a low cost to boost NER tasks, the models still suffer from the noise introduced by the imperfect rules. Therefore, denoising models have been proposed to allow the model to better tolerate the imperfect annotations created by rules. \citet{shang2018learning} proposed AutoNER that trains NER systems using only lexicons with a tie-or-break tagging denoising model. Similarly,
some recent work~\cite{liu2019towards,yang2018distantly,cao2019low} 
have used a partial matching module to denoise the noisily labeled data sets.


Recently, weak supervision is proposed as another form of denoising framework without using any labeled data. Weakly supervised systems use handcrafted labeling rules to create weak training instances from unlabeled data and then use a denoising generative model to approximate the posteriors of the rules. In this process, the unknown gold label is treated as latent variable by the generative model. The performance of the labeling rules could be further enhanced by training a neural network-based discriminative model by treating the posteriors as soft labels. Related studies such as Snorkel~\cite{ratner2020snorkel}, SwellShark~\cite{fries2017swellshark}, LinkedHMM~\cite{safranchik2020weakly}, and~\cite{lison2020named} have demonstrated great success with carefully curated labeling rules. 

However, manually designing those high-quality rules often require domain expertise and easy to have a low sensitivity on identifying entities. In~\cite{lison2020named}, the weak training data is created by broadly collecting available labeling rules from multiple sources, which demonstrates the importance of being able to automatically find new heuristics missed by human efforts. To find new heuristic rules on the basis of a relatively limited number of manually designed rules, previous studies have tried bootstrapping by relying on the co-occurrence, context and pattern features~\cite{thelen2002bootstrapping,riloff2003learning,yangarber2003counter,shen2017setexpan,tao2015leveraging,berger2018visual,yan2019learning}. 

Recent studies on graph neural networks has opened up another possibility for learning new rules. By internally infusing the semantics of the neighboring nodes, the popular Graph Convolutional Network (GCN)~\cite{kipf2016semi} and Graph Attention Network (GAT)~\cite{velivckovic2017graph} have shown great success in semi-supervised node classification when the number of labeled nodes is limited.  
Graph neural networks have been applied for many NLP tasks such as text classification~\cite{yao2019graph,zhang2019aspect,hu2019heterogeneous}, semantic role labeling~\cite{marcheggiani2017encoding}, 
machine translation~\cite{beck2018graph}, 
question answering~\cite{song2018exploring,saxena2020improving}, information extraction~\cite{liu2018jointly,vashishth2018reside,nguyen2018graph,sahu2019inter,sun2019joint,fu2019graphrel,zhang2019attention}, etc. In our work, we proposed to use graph neural networks to learn new labeling rules.
Based on Graph Attention Network~\cite{velivckovic2017graph}, we designed a new graph network model with a new loss function. Experimental results demonstrated that our model performed better than the original graph attention network on learning accurate labeling rules.

\section{Conclusion}
In this work, 
we proposed a weakly supervised NER framework that automatically learns high-quality new rules from only a handful of manually designed rules with  
a graph-based labeling rule augmentation method (\textsc{GLaRA}).
Experiments on three NER datasets demonstrate that our model outperforms baseline systems and achieves substantially better performance when the number of manual rules is limited. 
In addition, we also defined six types of rules that have been demonstrated useful for recognizing entities.
In the future, we plan to improve {\sc GLaRA} by investigating more complex rule types and rule representation methods for weakly supervised NER.

\bibliography{eacl2021}
\bibliographystyle{acl_natbib}

\appendix

\section{Hyperparameter configuration}
\label{tab:append-conf}
\paragraph{Graph Propagation Model} We use the same graph architecture to train propagation for all types of rules. The model contains $2$ graph attention layers each with $3$ attention heads and dropout rate is set to be $0.5$. The hidden size of each layer is set as $64$. We use Adam optimizer with a learning rate of $0.0001$. Other hyperparameters (training epoch and number of selected new rules) are presented in Table~\ref{tab:propagation-param}.
\begin{table}[h]
\small
\begin{tabular}{p{21mm}p{4mm}p{4mm}p{4mm}p{4mm}p{4mm}p{4mm}}\hline
\multirow{2}{*}{Hyperparam}  & \multicolumn{2}{c}{NCBI} & \multicolumn{2}{c}{BC5CDR} & \multicolumn{2}{c}{Laptop} \\\cline{2-7}
& epoch  & \#rules  & epoch  & \#rules   & epoch  & \#rules  \\\hline
SurfaceForm   & 50   & 50  & 50  & - & 50  & 25   \\
Prefix   & 50   & -  & 50     & - & 50 & 15   \\
Suffix    & 50   & 25  & 50     & 25(\mtt{D})   & 50  & 25 \\
PreNgram$_{\rm inclusive}$    & 50  & 25  & 30  & -& 30   &  \\
PreNgram$_{\rm exclusive}$        & 50 & - & 50     & 25(\mtt{D})   & 30  &  \\
PostNgram$_{\rm inclusive}$   & 50    & -& 30   &  & 50 &  \\
PostNgram$_{\rm exclusive}$         & 50 & -  & 50     & 30(\mtt{C})  & 50  &  \\
Dependency  & 50    & 25  & 50   &  - & 50 & 25    \\\hline      
\end{tabular}
\caption{Summary of hyperparameters for propagation. In BC5CDR data. ``\mtt{D}'' denotes the number of selected rules for Disease and ``\mtt{C}'' denotes the that for Chemical for the corresponding rule type. ``-'' means the corresponding type of propagated rules are not used on our final model.}
\label{tab:propagation-param}
\end{table}

\paragraph{Generative Model} Table~\ref{tab:generative-param} presents the hyperparameters used for tuning our LinkedHMM generative model, including ``Initial Accuracy (estimated initial accuracy of the rules)'', ``Accuracy Prior (regularization for initial accuracy)'', and ``Balance Prior (the entity class distribution)''. We used grid search to find the best hyperparameters. The search ranges are created around the default settings of the LinkedHMM model on the three data sets. For training epoch, we use the default setting, $5$, for all three datasets. For more details about the hyperparameters, please refer to~\cite{safranchik2020weakly}.

\begin{table}[h]
\small
\begin{tabular}{p{10mm}p{5mm}p{15mm}p{15mm}p{15mm}}\hline
\textbf{Hyperparam}  &  & NCBI & CDR  & Laptop \\\hline
\multirow{2}{*}{Init Acc} & Search & [0.75-0.95] & [0.75-0.95] & [0.75-0.95]\\
                                  & Best   & 0.85 & 0.85 & 0.9\\\hline
\multirow{2}{*}{Acc Prior} & Search & [45-65] & [0-15] & [0-5]\\
                                  & Best   & 55 & 5 & 1\\\hline
\multirow{2}{*}{Bal Prior} & Search & [440, 460] & [440, 460] & [0, 20]\\
                                  & Best   & 450 & 450 & 10\\\hline

\end{tabular}
\caption{Summary of hyperparameters for training the LinkedHMM generative model on each data set. The search steps for \mtt{Initial Accuracy}, \mtt{Accuracy Prior}, and \mtt{Balance Prior} are $0.05$, $5$, and $5$, respectively.}
\label{tab:generative-param}
\end{table}

\paragraph{Discriminative Model}Table~\ref{tab:discriminative-param} presents the hyperparameter configuration for training discriminative model (BiLSTM-CRF). BERT$_{base}$ is used to extract word embeddings and fine-tuned with BiLSTM-CRF. All discriminative models are trained on a 11G 1080Ti GPU with training time being up to $\sim$20s$/$epoch. All models have $\sim$110M parameters. 

\begin{table}[h]
\small
\begin{tabular}{l@{\hskip 2pt}l@{\hskip 2pt}l@{\hskip 2pt}l@{\hskip 2pt}l@{\hskip 2pt}l@{\hskip 2pt}}\hline
Hyperparam  &        & NCBI    & BC5CDR     & Laptop \\\hline
BERT   &   & SciBERT & SciBERT & BERT   \\
\multirow{2}{*}{BiLSTM} &Hidden Dim & 256  & 256     & 256    \\
                         & Dropout       & 0.1     & 0.1     & 0.1    \\
CRF & &  yes     & yes       & no    \\\hline
\multirow{2}{*}{AdamW}    & Learning Rate & 1e-4   & 1e-4  & 1e-4  \\
                         & Epoch  & 30 & 30 & 30     \\\hline
Batch size & & 8 & 8 & 8\\\hline
Max sent length & & 128 & 128 & 128\\\hline
\end{tabular}
\caption{Summary of hyperparameters for training discriminative model on each data set. }
\label{tab:discriminative-param}
\end{table}

\section{Performance on Development Data}
In Table~\ref{tab:dev_results}, we present the performance of the LinkedHMM model on developement sets, with the additional seeding rules manually selected by us, referred as LinkedHMM-M, the performance of the following discriminative model, referred as LinkedHMM-M-D. Also, we report the performance of our models, {\sc GLaRA} and {\sc GLaRA-D} (with discriminative model) on development sets.

\begin{table}[h]
\small
\begin{tabular}{lccc}
\toprule
{\bf Model} & {\bf NCBI} &{\bf BC5CDR} & {\bf Laptop} \\
\midrule
LinkedHMM-M & 82.3 & 87.5& 70.1\\
LinkedHMM-M-D & 82.8{\tiny$\pm$.3} & 84.5{\tiny$\pm$.2} & 71.5{\tiny$\pm$.8}\\
\textsc{GLaRA} & 83.1{\tiny $\pm$.2} &87.4{\tiny$\pm.3$}  & 72.3{\tiny$\pm.8$}\\ 
\textsc{GLaRA-D}& 83.4{\tiny$\pm$.3} &  84.5{\tiny$\pm.1$} & 72.6{\tiny$\pm.6$} \\
\bottomrule
\end{tabular}
\caption{Micro F1 performance on each development data set. \mtt{LinkedHMM-M} denotes the baseline LinkedHMM model trained with the extra manually selected seeding rules. \mtt{LinkedHMM-M-D} denotes the discriminative mode (LSTM-CRF) trained based on \mtt{LinkedHMM-M}. Similarly, {\sc GLaRA}\mtt{-D} denotes the LSTM-CRF model trained based on {\sc GLaRA}.
}
\label{tab:dev_results}
\end{table}

\section{Manually Selected Seeds for propagation}

As mentioned in the paper, the baseline LinkedHMM does not have seeding rules for all types of rules. For example, negative seeding rules are missing from the baseline LinkedHMM model, except that \mtt{SurfaceForm} rules uses a list of stopwords as negative seeding rules. For some types of seeding rules, there are only a few available that are not good enough for training the propagation model. Therefore, for the rule types that do not have enough seeds, we manually select a small set of additional rules as seeds. Note that we keep the total number of seeding rules (including the ones from baseline system) less than $15$. We report the manually selected seeding rules for NCBI, BC5CDR-Disease, BC5CDR-Chemical, and LaptopReview in Table~\ref{tab:seeds-ncbi}, Table~\ref{tab:seeds-cdr-dis}, Table~\ref{tab:seeds-cdr-chem}, and Table~\ref{tab:seeds-latop}, respectively.

\begin{table*}[t]
\vspace{-0.5cm}
\centering
\small
\begin{tabular}{lll}
\toprule
Rules     &     & NCBI  \\\hline
\mtt{Surface}   & pos & - \\
          & neg & -\\\hline
\mtt{Suffix}    & pos & *skott, *drich, *umour, *axia, *iridia \\
          & neg & *ness, *nant, *tion, *ting, *enesis, *riant, *tein, *sion, *osis, *lity\\
\mtt{Prefix}    & pos & carc*, myot*, tela*, ovari*,atax*, carcin*, dystro*\\
          & neg & defi*, comp*, fami*, poly*, chro*, prot*, enzym*, sever*, develo*, varian* \\\hline
\mtt{exclusive~PreNgram}  & pos & suffer\_from\_*, fraction\_of\_*, pathogenesis\_of\_*, cause\_severe\_*\\
          & neg & -pron\_*, suggest\_that\_*, -\_cell\_*, presence\_of\_*, expression\_of\_*, majority\_of\_*\\
          &     & loss\_of\_*, associated\_with\_*,impair\_in\_*, cause\_of\_*, defect\_in\_*, family\_with\_*\\\hline
\mtt{inclusive~PreNgram}  & pos & breast\_and\_ovarian\_*, x\_-\_link\_*, breast\_and\_*, stage\_iii\_*, myotonic\_* \\
          & neg & enzyme\_*, primary\_*, non\_-\_*, \\\hline
\mtt{exclusive~PostNgram} & pos &  \\
          & neg & *\_and\_the, *\_cell\_line, *\_in\_the \\\hline
\mtt{inclusive~PostNgram} & pos & *\_-\_t, *\_cell\_carcinoma, *\_muscular\_dystrophy, *\_'s\_disease, *\_carcinoma, *\_dystrophy \\
          & neg & *\_muscle, *\_ataxia, *\_'system, *\_defect , *\_other\_cancer, *\_i, *\_ii\\\hline
\mtt{DepRule}       & pos & 

 FirstTokenDep:amod\_LastToken:dystrophy, FirstTokenDep:punct\_LastToken:telangiectasia,\\
&     &  FirstTokenDep:compound\_HeadSurf:t, FirstTokenDep:amod\_LastToken:dysplasia\\
&     &    SecondLastTokenDep:compound\_LastToken:syndrome,\\
& neg & FirstTokenDep:amod\_LastToken:deficienc, FirstTokenDep:amod\_LastToken:deficiency\\
  &     & FirstTokenDep:amod\_LastToken:defect, FirstTokenDep:pobj\_LastToken:cancer\\
   &     & SecondLastTokenDep:compound\_LastToken:cancer, \\
   &     &SecondLastTokenDep:compound\_LastToken:disease\\
    &     &       SecondLastTokenDep:appos\_LastToken:t, SecondLastTokenDep:compound\_LastToken:t \\

\bottomrule
\end{tabular}
\caption{Manually selected seeding rules for NCBI dataset. ``-'' means no seeding rules selected, and we only use the rules provided in baseline.}
\label{tab:seeds-ncbi}
\end{table*}

\begin{table*}[t]
\centering
\small
\begin{tabular}{lll}
\toprule
Rules     &     & BC5CDR (Disease)  \\\hline
\mtt{Surface}   & pos & - \\
          & neg & -\\\hline
\mtt{Suffix}    & pos & *epsy, *nson \\
          & neg & *ing, *tion, *tive, *lity, *mone, *fect, *crease, *sion, *lion,\\
          &     & *elet, *gical, *nosis, *sive, *ment, *tory, *sionetic, *ency, *ture, \\\hline
\mtt{Prefix}    & pos & anemi*, dyski*, heada*, hypok*, hypert*, ische*, arthr*, hypox*, \\
          &     & toxic*, arrhyt*, ischem*, hypert*, dysfunc*\\
          & neg & symp*, resp*, funct*, inter*, decre*, prote*, neuro*, cardi*, myoca*, ventr*, decre*\\
          &     & syst* \\\hline
\mtt{exclusive~PreNgram}  & pos & to\_induce\_*, w\_-\_*, and\_severe\_*, suspicion\_of\_*, die\_of\_*, have\_severe\_*, \\
          &     & of\_persistent\_*, cyclophosphamide\_associate\_*\\
          & neg &seizure\_and\_*, symptom\_and\_*, dysfunction\_and\_*, failure\_with\_*, sign\_of\_*, lead\_to\_*\\\hline
\mtt{inclusive~PreNgram}  & pos & parkinson\_'s\_*, torsade\_de\_*, acute\_liver\_*, neuroleptic\_*, malignant\_*, alzheimer\_'s\_*, \\
          &     & congestive\_heart\_*, migraine\_with\_*, sexual\_side\_*, renal\_cell\_*, tic\_-\_* \\
          & neg & renal\_function\_*, decrease\_in\_*, increase\_in\_*, reduction\_in\_*, rise\_in\_*, loss\_of\_*\\
          &     & chronic\_liver\_*, abnormality\_in\_*, human\_immunodeficiency\_*, optic\_nerve\_*, drug\_-\_*\\
          &     & non\_-\_* \\\hline
\mtt{exclusive~PostNgram} & pos & - \\
          & neg & *\_and\_the, *\_cell\_line, *\_in\_the \\\hline
\mtt{inclusive~PostNgram} & pos & *\_'s\_disease, *\_infarction, *\_'s\_sarcoma, *\_epilepticus , *\_artery\_disease, *\_de\_pointe\\
          &     &  *\_insufficiency, *\_with\_aura, *\_artery\_spasm, *\_'s\_encephalopathy \\
          & neg & *\_toxicity, *\_pain, *\_fever, *\_function, *\_blood\_pressure, *\_effect, *\_impairment, *\_loss\\
          &     &  *\_event, *\_protein, *\_pressure, *\_impair, *\_phenomenon, *\_system, *\_side\_effect\\
          &     & *\_of\_disease\\\hline
\mtt{DepRule}       & pos &

FirstTokenDep:poss\_LastToken:disease\\
&     & FirstTokenDep:compound\_LastToken:cancer, FirstTokenDep:amod\_LastToken:dysfunction\\
&     & FirstTokenDep:compound\_LastToken:disease, FirstTokenDep:compound\_LastToken:failure\\
&     & FirstTokenDep:compound\_LastToken:anemia, FirstTokenDep:compound\_LastToken:cancer\\
&     & SecondLastTokenDep:pobj\_LastToken:disease\\

          & neg & 
FirstTokenDep:amod\_LastToken:toxicity\\
&     & FirstTokenDep:amod\_LastToken:impairment, FirstTokenDep:amod\_LastToken:syndrome\\
&     & FirstTokenDep:amod\_LastToken:complication, FirstTokenDep:amod\_LastToken:symptom\\
&     & FirstTokenDep:amod\_LastToken:damage, FirstTokenDep:amod\_LastToken:disease\\
&     & FirstTokenDep:amod\_LastToken:function, FirstTokenDep:amod\_LastToken:damage\\
&     & FirstTokenDep:compound\_LastToken:loss, SecondLastTokenDep:nmod\_LastToken:b\\
&     & SecondLastTokenDep:conj\_LastToken:arrhythmia\\ 
&     & SecondLastTokenDep:pobj\_LastToken:symptom\\

\bottomrule
\end{tabular}
\caption{Manually selected seeding rules for BC5CDR (Disease) dataset. ``-'' means no seeding rules selected, and we only use the rules provided in baseline.}
\label{tab:seeds-cdr-dis}
\end{table*}

\begin{table*}[]
\small
\centering
\begin{tabular}{lll}
\toprule
Rules     &     & BC5CDR (Chemical)  \\\hline
\mtt{Surface}   & pos & - \\
          & neg & -\\\hline
\mtt{Suffix}    & pos & *pine, *icin, *dine, *ridol, *athy, *zure, *mide, *fen, *phine \\
          & neg & *ing, *tion, *tive, *tory, *inal, *ance, *duce, *atory, *mine, *line, *tin,\\
          &     & *rate, *late, *ular, *etic, *onic, *ment, *nary, *lion, *ysis, *logue, *mone \\\hline
\mtt{Prefix}    & pos & chlor*, levo*, doxor*, lithi*, morphi*, hepari*, ketam*, potas* \\
          & neg & meth*, hepa*, prop*, contr*, pheno*, contra*, acetyl*, dopami* \\\hline
\mtt{exclusive~PreNgram}  & pos & dosage\_of\_*, sedation\_with\_*, mg\_of\_*, application\_of\_*, -\_release\_*, ingestion\_of\_*\\
          &     & intake\_of\_*\\
          & neg & of\_to\_*, to\_the\_*, be\_the\_*, with\_the\_*, in\_the\_*, on\_the\_*, for\_the\_*\\\hline
\mtt{inclusive~PreNgram}  & pos & external\_*, mk\_*, mk\_-\_*, cis\_*, cis\_-\_*, nik\_*, nik\_-\_*, ly\_*, ly\_-\_*, puromycin\_*\\
          & neg & reduce\_*, all\_*, a\_*, the\_*, of\_*, alpha\_*, alpha\_-\_*, beta\_*, beta\_-\_*\\\hline
\mtt{exclusive~PostNgram} & pos &  *\_-\_associate, *\_-\_induced\\
          & neg & *\_-\_related, *\_that\_of, *\_of\_the \\\hline
\mtt{inclusive~PostNgram} & pos & *\_aminocaproic\_acid, *\_-\_aminocaproic *\_acid, *\_retinoic\_acid, *\_dopa, *\_tc\\
          &     & *\_-\_aminopyridine, *\_aminopyridine, *\_-\_penicillamine, *\_-\_dopa, *\_-\_aspartate, *\_fu\\ 
          &     & *\_hydrochloride \\
          & neg & *\_drug, *\_cocaine, *\_calcium, *\_receptor\_agonist, *\_blockers, *\_block\_agent\\
          &     & *\_inflammatory\_drug\\\hline 
\mtt{DepRule}       & pos & 
FirstTokenDep:amod\_LastToken:oxide, FirstTokenDep:compound\_LastToken:chloride\\
&     & FirstTokenDep:amod\_LastToken:acid, FirstTokenDep:compound\_LastToken:acid \\
&     & FirstTokenDep:compound\_LastToken:hydrochloride\\
&     & SecondLastTokenDep:amod\_LastToken:aminonucleoside,\\

& neg & 
FirstTokenDep:compound\_LastToken:a, SecondLastTokenDep:pobj\_LastToken:acid\\ 
&     & SecondLastTokenDep:pobj\_LastToken:a, SecondLastTokenDep:pobj\_LastToken:the\\
&     & SecondLastTokenDep:pobj\_LastToken:a\\
\bottomrule
\end{tabular}
\caption{Manually selected seeding rules for BC5CDR (Chemical) dataset. ``-'' means no seeding rules selected, and we only use the rules provided in baseline.}
\label{tab:seeds-cdr-chem}
\end{table*}

\begin{table*}[h]
\small
\centering
\begin{tabular}{lll}
\toprule
Rules     &     & LatopReview  \\\hline
\mtt{Surface}   & pos & - \\
          & neg & -\\\hline
\mtt{Suffix}    & pos & *pad, *oto, *fox, *chpad, *rams \\
          & neg & *ion, *ness, *nant, *lly, *ary\\\hline
\mtt{Prefix}    & pos & feat*, softw*, batt*, Win*, osx* \\
          & neg & pro*, edit*, repa*, rep*, con*, dis*, appl*, equip* \\\hline
\mtt{exclusive~PreNgram}  & pos &  - \\
          & neg & in\_the\_*, on\_the\_*, for\_the\_*, -pron\_*\\\hline
\mtt{inclusive~PreNgram}  & pos & windows\_*, hard\_*, extended\_*, touch\_*, boot\_*\\
          & neg & mac\_*, apple\_*, a\_*, launch\_*, software\_* \\\hline
\mtt{exclusive~PostNgram} & pos & *\_and\_seal, *\_that\_come\_with\\
          & neg & *\_shut\_down, *\_do\_not\_work,  \\\hline
\mtt{inclusive~PostNgram} & pos & *\_x, *\_xp, *\_vista, *\_drive, *\_processing\\
          & neg & *\_screen, *\_software, *\_quality, *\_technical, *\_cut\\\hline
\mtt{DepRule}       & pos &
FirstTokenDep:compound\_LastToken:port, FirstTokenDep:compound\_LastToken:button\\
&     & FirstTokenDep:nummod\_LastToken:ram, FirstTokenDep:amod\_LastToken:drive\\
& neg & 
FirstTokenDep:compound\_LastToken:option, SecondLastTokenDep:pobj\_LastToken:plan\\
&     & SecondLastTokenDep:nsubj\_LastToken:design, SecondLastTokenDep:pobj\_LastToken:plan\\
\bottomrule
\end{tabular}
\caption{Manually selected seeding rules for LaptopReview dataset. ``-'' means no seeding rules selected, and we only use the rules provided in baseline.}
\label{tab:seeds-latop}
\end{table*}

\end{document}